\documentclass[authoryear,preprint,review,12pt]{elsarticle}

\usepackage{epsfig}

\usepackage{amssymb}
\usepackage{amsmath}
\usepackage{amsthm}

\usepackage{graphicx}
\usepackage{subcaption}
\usepackage{caption}

\usepackage{booktabs}
\usepackage{adjustbox}
\usepackage{float}
\newpageafter{abstract}
\journal{Computers and Electronics in Agriculture}

\begin{document}

\begin{frontmatter}



\title{Unlocking Zero-Shot Plant Segmentation with \\ Pl@ntNet Intelligence}


\author[1]{Simon Ravé}
\author[2]{Jean-Christophe Lombardo}
\author[1]{Pejman Rasti}
\author[2]{Alexis Joly}
\author[1]{David Rouseau}

\affiliation[1]{organization={University of Angers, LARIS, IRHS, INRAe}, 
            country={France}}

\affiliation[2]{organization={Inria}, city={Montpellier}, country={France}}
\begin{abstract}
We present a zero-shot segmentation approach for agricultural imagery that leverages Plantnet, a large-scale plant classification model, in conjunction with its DinoV2 backbone and the Segment Anything Model (SAM). Rather than collecting and annotating new datasets, our method exploits Plantnet’s specialized plant representations to identify plant regions and produce coarse segmentation masks. These masks are then refined by SAM to yield detailed segmentations. We evaluate on four publicly available datasets of various complexity in terms of contrast including some where the limited size of the training data and complex field conditions often hinder purely supervised methods. Our results show consistent performance gains when using Plantnet-fine-tuned DinoV2 over the base DinoV2 model, as measured by the Jaccard Index (IoU). These findings highlight the potential of combining foundation models with specialized plant-centric models to alleviate the annotation bottleneck and enable effective segmentation in diverse agricultural scenarios.
\end{abstract}







\end{frontmatter}




\section{Introduction}

Computer vision has become crucial in agricultural tasks, where standardizing plant observations, enhancing productivity, and extracting features hard to detect for the human eye are crucial \citep{mochidaComputerVisionbasedPhenotyping2018, LI2020105672}. However, plant diversity, complex field backgrounds, and unpredictable environmental conditions pose significant challenges for vision-based approaches. Deep learning architectures, notably convolutional neural networks \citep{6795724, pound2016}, have demonstrated promising performance in automating feature extraction, but they typically require large amounts of annotated data \citep{dlagri2018, PATRICIO201869} or complex features obtained using hyperspectral imaging or depth informations \citep{DEVANNA2025109611, SAHIN2023107956}. Acquiring such labeled data is often laborious in plant-focused scenarios with limited variability, shifting the bottleneck to data collection and labeling. Some 

A more recent development involves foundation models \citep{Bommasani2021OnTO, DBLP:journals/corr/abs-2103-00020}, which are trained on massive datasets and can be adapted to various downstream tasks with minimal supervision. In agriculture, leveraging generalist foundation models has emerged as a viable approach \citep{Chen_2023_ICCV, ZHAO2023103946}, but it remains suboptimal when these models lack explicit plant knowledge. One candidate to address this gap is Plantnet \citep{barthelemy:hal-02810776, 10.1145/2502081.2502251}, a large-scale crowd-sourced database of more than 50,000 plant species. Plantnet’s model, a fine-tuned version of DinoV2 \citep{oquab2023dinov2}, builds on self-supervised vision transformers \citep{dosovitskiyImageWorth16x162020} to provide robust plant-specific features. While Plantnet is primarily used for species identification \citep{JOLY201422, pitman:hal-03312029, hoye:hal-04405026, elvekjaer:hal-04593804}, its potential for other tasks, such as semantic segmentation or soil coverage estimation, remains largely unexplored.

Concurrently, open-set segmentation has gained attraction with models like Segment Anything Model (SAM) \citep{kirillov2023segment}, which promises broad applicability including in agricultural scenarios \citep{SAEIDIFAR2024109436, FERREIRA2025110086}. Yet SAM’s performance on agricultural imagery has been merely satisfactory, often misidentifying small crops due to a limited agricultural training corpus \citep{samNotPerfect}. Combining SAM with plant-specific knowledge could yield more accurate segmentation of complex canopies and subtle plant features.

Given the scarcity of high-quality annotated plant datasets, evaluating new methods often relies on limited or localized data. To address this, we employ various open-source datasets, including Phenobench, where state-of-the-art supervised models already achieve high Jaccard scores, and others that represent diverse field conditions. In this article, we propose a zero-shot plant segmentation approach that fuses Plantnet and generalist foundation models. We compare our method against a supervised baseline, and its scaling law for the number of training samples, demonstrating the feasibility of using Plantnet’s specialized representations for soil coverage estimation and plant semantic segmentation without extensive manual annotation.

\section{Material and methods}

\subsection{Preprocessing} \label{preprocessing}

When using the Plantnet Model, we need to preprocess our images to the required format. Because Plantnet uses DinoV2, a Vision Transformer with absolute positionnal embedding, we can interpolate the positionnal embedding. This allows us to process images of varying sizes without needing to crop them or resize them, at the cost of lot more computes. Nonetheless, due to compute limitation we limit our images to 1036 pixels on the smallest edge. 1036 is chosen because it is double the training size of the base DinoV2 model and is a multiple of 14, the size of the models tokens patches. If the image is not square, we pad the other edges to the nearest multiple of 14. This method allows us to keep more information on the image than by resizing and cropping it to the initial 518x518 pixels.

\subsection{Segmentation Approach}

\begin{figure}[ht]
    \centering
    \includegraphics[width=\linewidth]{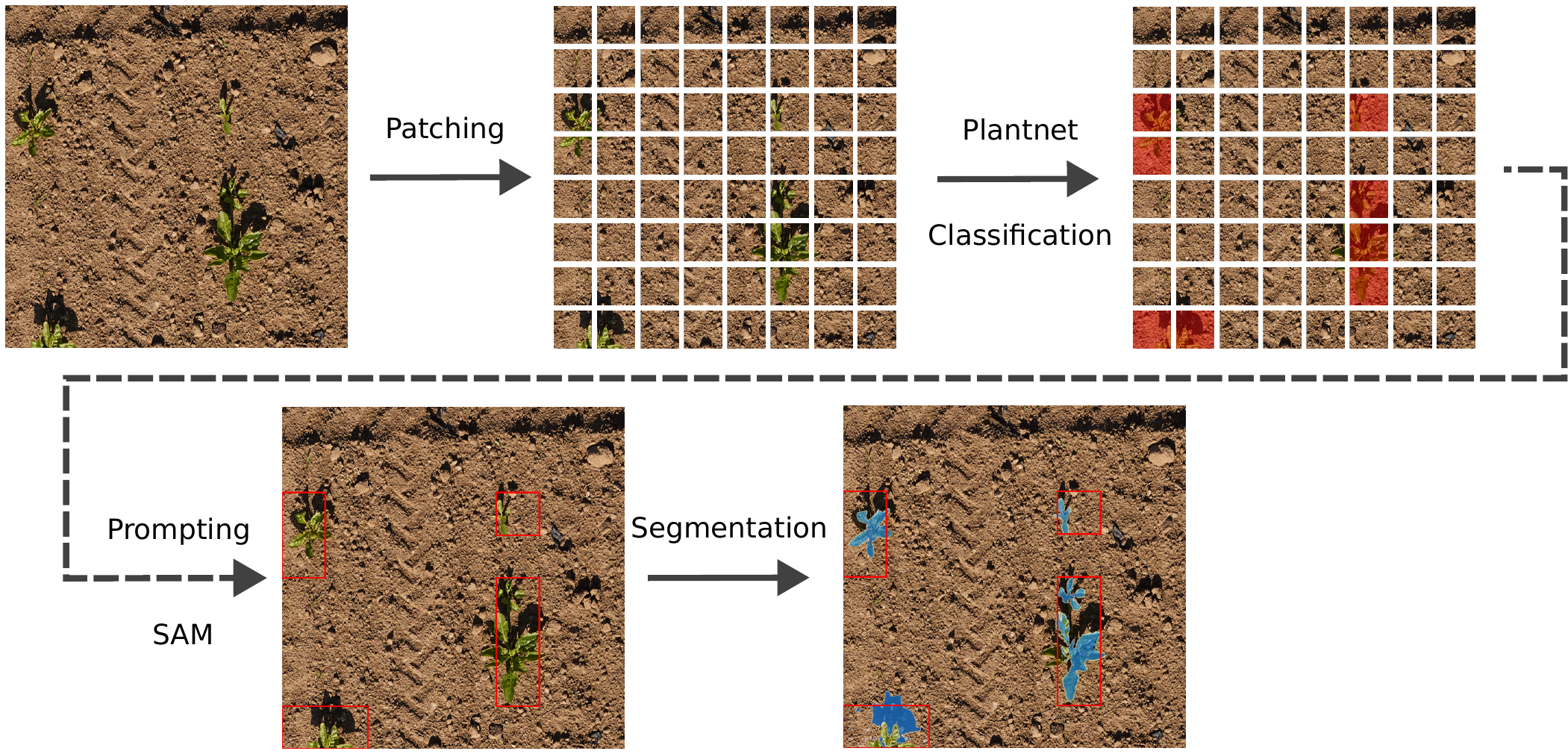}
    \caption{Method Pipeline, the image is patched and goes through plantnet, then each patch is classified creating a rough mask that is then turned into a box prompt and refined using SAM.}
    \label{fig:pipeline}
\end{figure}

Plantnet is a species classification model trained on 21 millions crowd-sourced images to classify more than 50,000 plant species \citep{Lefort2024}. We aim to determine whether we could leverage the existing knowledge of plantnet about plants and transfer it to another task: segmentation. Specifically, we want to segment plants from the background to isolate the plant of interest. We propose using Plantnet as an encoder to extract features, which are then aggregated in a zero-shot manner, without any retraining. These aggregated features are finally used as prompts for the Segment Anything Model 2 (SAM2).

\begin{figure}[ht!]
    \centering
    \includegraphics[width=\linewidth]{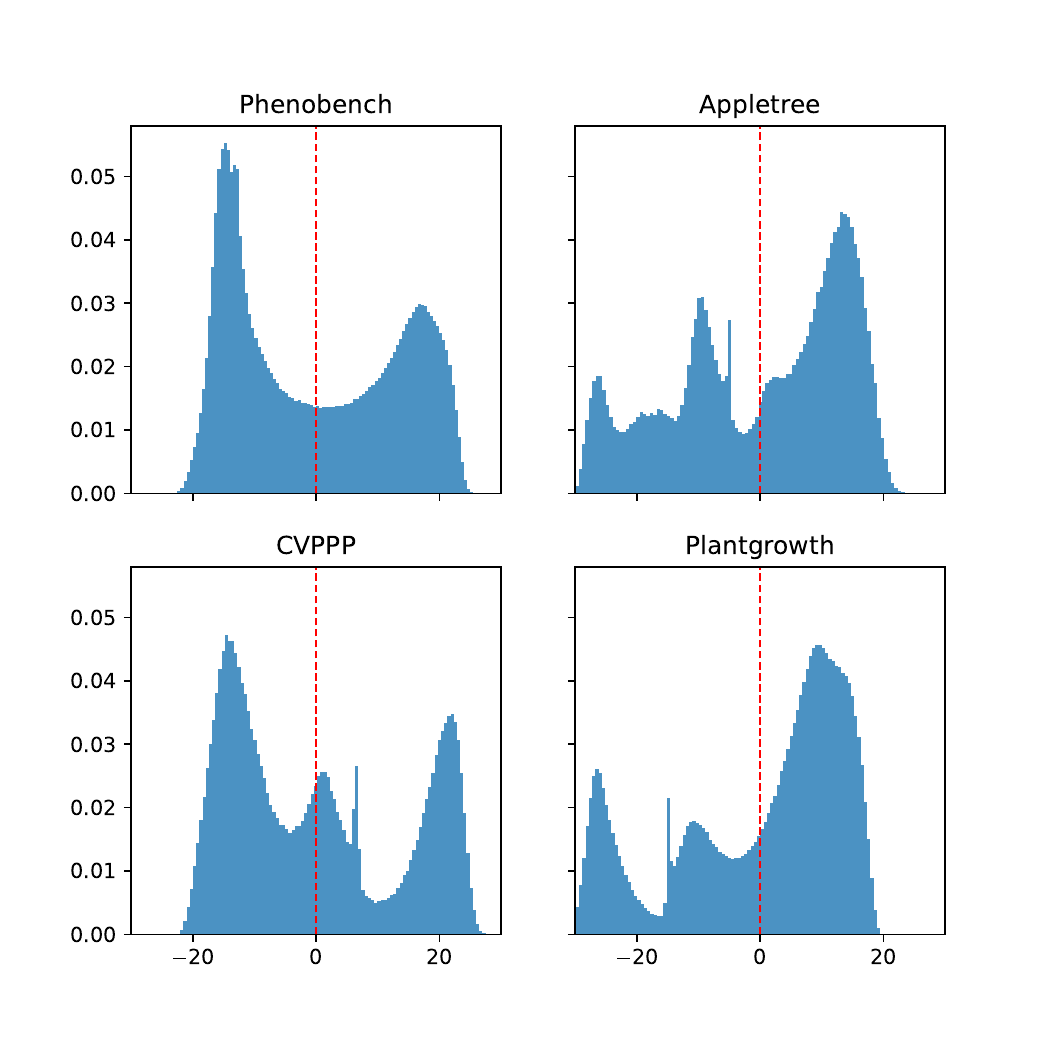}
    \caption{Analysis of the first PCA components of the output tokens from the Plantnet model on the four datasets. We observe a clear separation between positive and negative tokens. On Phenobench, 0 is clearly a local minimum. On the other datasets, although 0 is not a local minimum, it can still serve as an effective threshold for separating the tokens into two clusters.}
    \label{fig:tokens-pca}
\end{figure}

Using the Plantnet model backbone, which utilizes a Vision Transformer (ViT) DinoV2 architecture \citep{oquab2023dinov2}, we extract the output token features from all images. Then we compute a Principal Component Analysis (PCA) over the token features extracted from the entire validation set of the current dataset, which means the PCA is computed using similar images. Subsequently, we classify each token as representing either plant or background by thresholding the first principal component at zero, where values $\geq 0$ indicate plant regions and values $< 0$ indicate background. Consistent with \citep{oquab2023dinov2}, thresholding at zero has been empirically confirmed as optimal for generalization across various datasets, as shown in Figure \ref{fig:tokens-pca}. Attempting dataset-specific optimization of this threshold typically yields negligible improvements while introducing an additional hyperparameter, reinforcing that the first PCA component inherently captures the primary subject of interest—in our case, plants. Ultimately, these classified tokens can be grouped to generate bounding box prompts or directly serve as preliminary masks. When resized to 256x256 pixels, these masks can subsequently be refined using Segment Anything Model 2 (SAM2).

To compare our method, and assess the added value of using Plantnet, we tested the same method but using the base DinoV2 model, which has not been fine tuned on plants data. DinoV2 is pretrained on the proprietary LVD-124M dataset from Meta, that is not specialized for plants.

\subsection{Datasets}



To evaluate our methods, we employed several datasets encompassing different viewing angles and varying plant densities, as detailed in Table \ref{table:datasets-examples}. The first dataset, Phenobench \citep{weyler2023pami}, comprises top-view images of growing sugar beet plants. Initially designed for weed detection among beet crops, we adapted this dataset by merging all individual plant masks into a simplified semantic segmentation task targeting every plants. We categorize Phenobench as a "sparse" dataset since the plants are distinctly separated, resulting in minimal occlusion or interference between adjacent plants. Such sparsity typically simplifies model training in supervised learning scenarios due to the clear constrast between plants and the background.

Additionally, we evaluated our methods using the Apple Tree Dataset \citep{agriculture13112097}. This dataset consists of profile images of apple trees captured outdoors, intended originally for isolating the primary apple tree positioned centrally within each image. Contrary to Phenobench, the Apple Tree Dataset exhibits significant plant overlap, classifying it as a "dense" dataset. This high density introduces substantial challenges for segmentation tasks, as overlapping trees complicate the accurate separation of individual trees, even under supervised conditions. Furthermore, the dataset contains a limited number of samples (150 images) with minimal variability, further increasing the difficulty of training robust segmentation models.

We also utilized two indoor plant datasets: the Plant Growth dataset \citep{PlantGrowth} and the CVPPP2017 dataset \citep{bell_2016_168158}. The latter, initially created for leaf-counting tasks, was repurposed for semantic segmentation by combining individual leaf masks into unified plant masks. Incorporating these datasets allowed us to construct diverse experimental scenarios encompassing sparse, dense, indoor, and outdoor plant segmentation contexts. This broad range of conditions enabled thorough and rigorous evaluation of our segmentation methods, ensuring their robustness and applicability in practical, real-world settings.

\begin{table}[ht]
\centering
\resizebox{\textwidth}{!}{
    \begin{tabular}{cccc}
    \toprule
    \textbf{Phenobench} & \textbf{AppleTreeDataset} & \textbf{Plant Growth} & \textbf{CVPPP2017}\citep{Minervini2015PRL} \\
    \midrule
    \includegraphics[height=4cm]{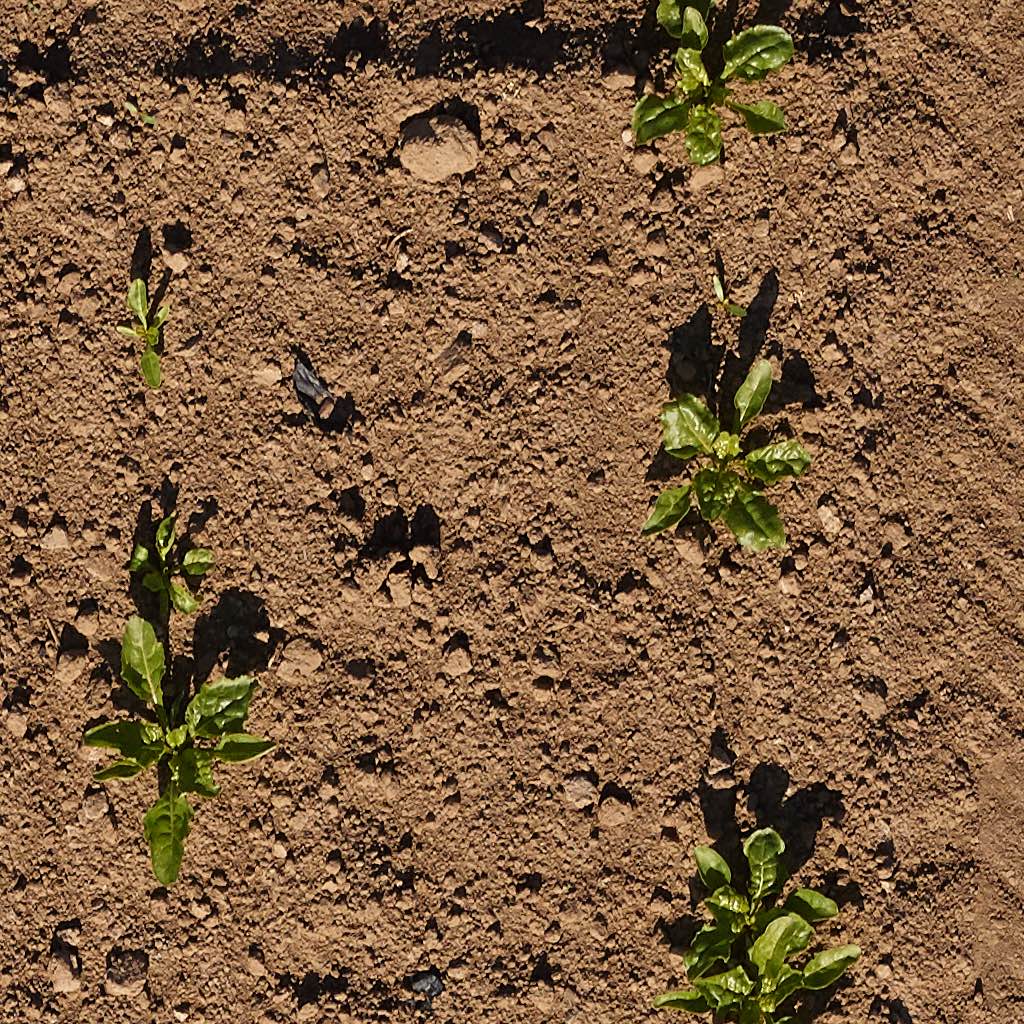} & 
    \includegraphics[height=4cm]{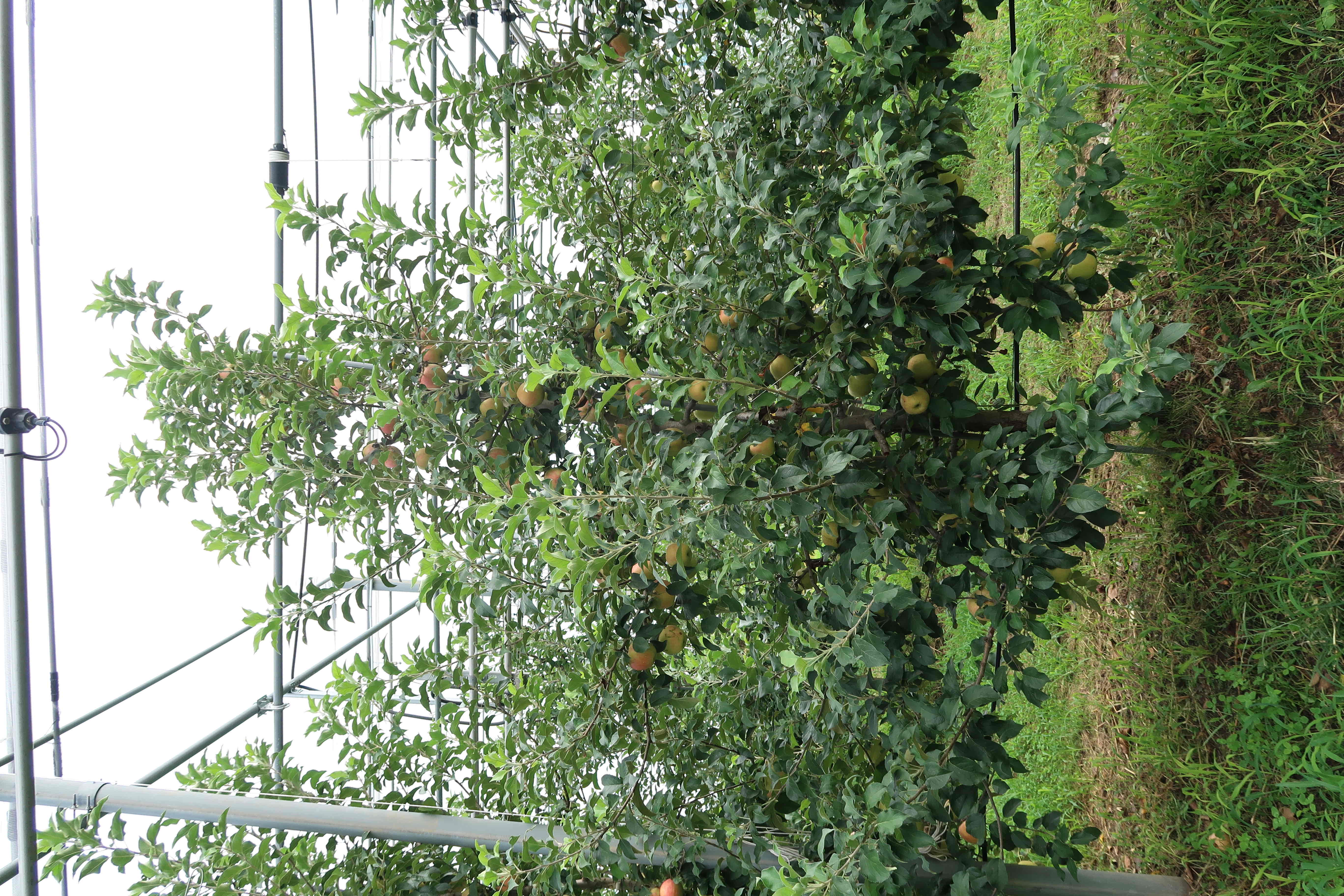} & 
    \includegraphics[height=4cm]{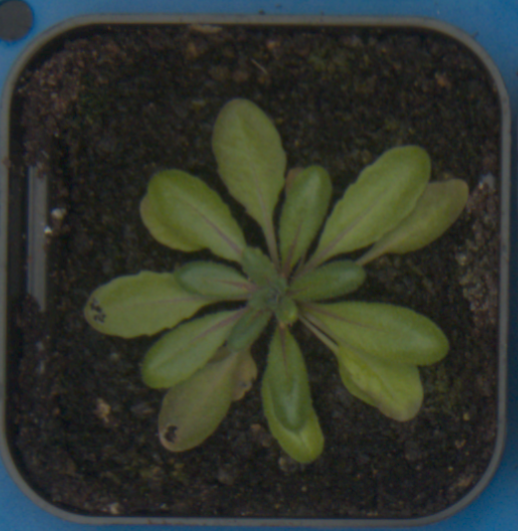} & 
    \includegraphics[height=4cm]{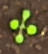} \\
    Sparse Outdoor & Dense Outdoor & Sparse Indoor & Sparse Indoor \\
    772 images & 150 images & 2008 images & 624 images \\
    \bottomrule
    \end{tabular}
}
\vspace{0.5cm}
\caption{Descriptions of datasets used.}
\label{table:datasets-examples}
\end{table}

\subsection{Evaluation}

To evaluate our method we rely on the Intersection Over Union (IoU) metric,
$$J(A,B) = \dfrac{\lvert A \cup B \lvert}{\lvert A \cap B\lvert}$$ which computes the ratio between the intersection of the predicted mask and the ground truth one against their union. When the IoU equals $1$, it means the prediction is perfectly on par with the true mask, and when it equals $0$ it means their is no overlap between the two

\section{Result}

As baseline, we compared our method that uses the Plantnet model to one using only the base DinoV2 weights \citep{oquab2023dinov2}. Table \ref{table:semantic-seg-results} highlights the significant  and systematic (for all 4 tested dataset) performance gain brought by the use of Plantnet compared to the baseline model.

\begin{table}[ht]
\centering
	\resizebox{\textwidth}{!}{
    \begin{tabular}{lccccc}
    \toprule
        \textbf{Model} & \textbf{Phenobench} & \textbf{AppleTreeDataset} & \textbf{Plant Growth} & \textbf{CVPPP2017}\citep{Minervini2015PRL}\\
        \midrule
        \textbf{DinoV2-Plantnet} & $\mathbf{0.672 \pm 0.289}$ & $\mathbf{0.714 \pm 0.120}$ &  $\mathbf{0.715 \pm 0.270}$ & $\mathbf{0.598 \pm 0.299}$ \\
        \textbf{DinoV2} & $0.119 \pm 0.171$ & $0.049 \pm 0.063$ &  $0.627 \pm 0.263$ & $0.466 \pm 0.358$ \\
        \bottomrule
    \end{tabular}
    }
\vspace{0.5cm}
\caption{Comparison of IoU metrics on multiples datasets using our method with either the base DinoV2 model or the DinoV2 model trained on the Plantnet Dataset. We see consistent improvements by using the model that was trained on plants.}
\label{table:semantic-seg-results}
\end{table}

Next, we wanted to see the impact of giving the rough mask made by thresholding the PCA first component to SAM in addition to the box prompt. The results in table \ref{table:sam-mask-input} shows that it does not improve the performance of the method except for the dense Apple Tree Dataset.

\begin{table}[ht]
\centering
	\resizebox{\textwidth}{!}{
    \begin{tabular}{lccccc}
    \toprule
        \textbf{Model} & \textbf{Phenobench} & \textbf{AppleTreeDataset} & \textbf{Plant Growth} & \textbf{CVPPP2017}\citep{Minervini2015PRL}\\
        \midrule
        \textbf{Without mask input} & $\mathbf{0.672 \pm 0.289}$ & $0.714 \pm 0.120$ & $\mathbf{0.715 \pm 0.270}$ & $\mathbf{0.598 \pm 0.299}$ \\
        \textbf{With masks input } & $0.651 \pm 0.283$ & $\mathbf{0.754 \pm 0.085}$ & $0.619 \pm 0.315$ & $0.590 \pm 0.288$ \\
        \bottomrule
    \end{tabular}
    }
\vspace{0.5cm}
\caption{Comparing our method with and without giving the rough 256x256 mask to SAM. Giving the masks to SAM seems to not improve the results, and even worsen them sometimes.}
\label{table:sam-mask-input}
\end{table}

\begin{figure}[ht!]
    \centering
    \includegraphics[width=\linewidth]{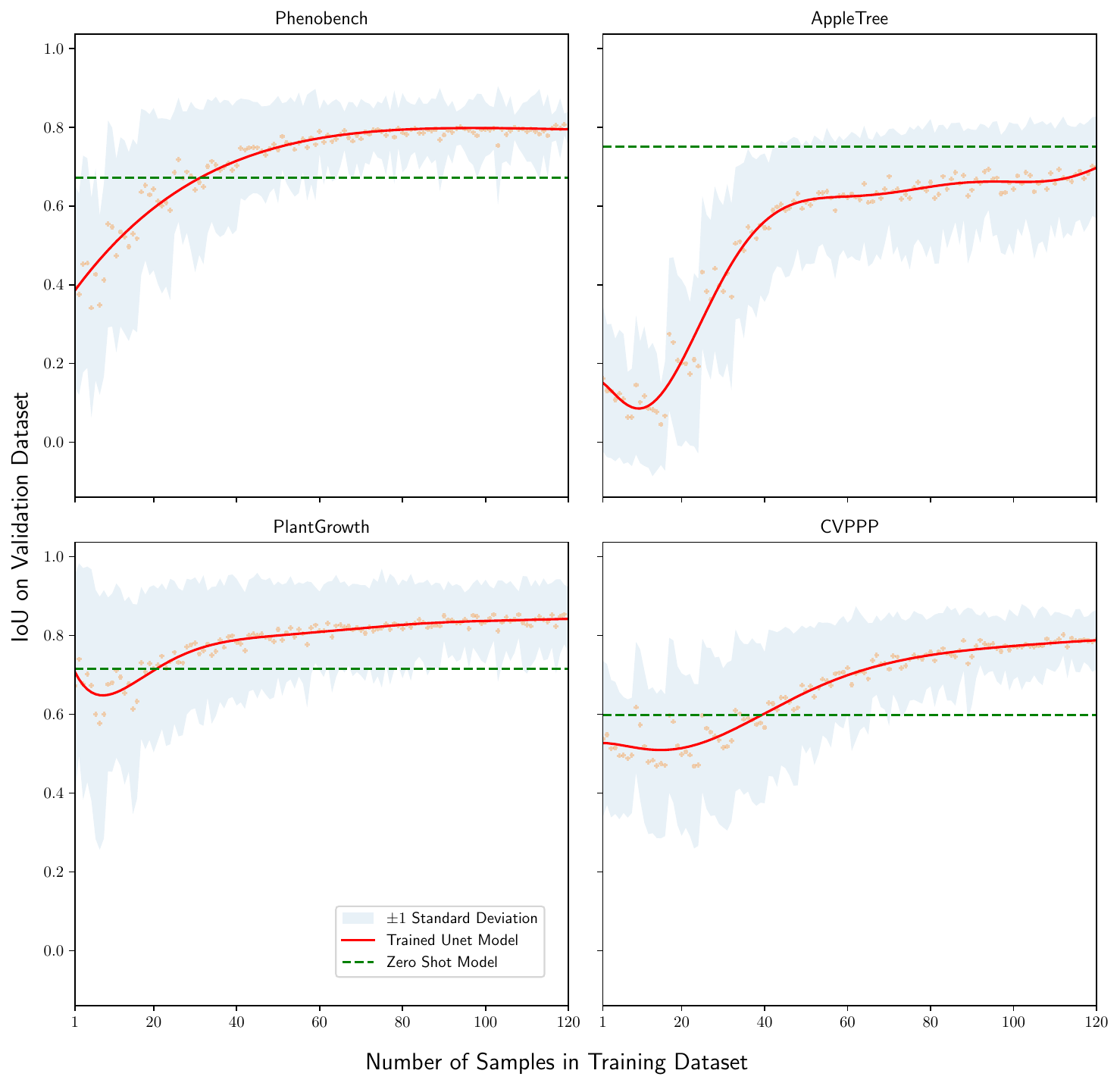}
    \caption{Evolution of U-Net Model Performance with increasing training data on 4 datasets. U-Net starts to outperform our method between 20 and 40 training samples for all datasets except the Apple Tree dataset where our method is always better. For each datasets size we did 100 differents training runs and computed the mean IoU on the validation datasets which are plotted in orange points, the blue envelope being the standards deviation of the models performances. }
    \label{fig:unet-evolution}
\end{figure}

In Figure \ref{fig:unet-evolution}, we compare the performance of a U-Net \citep{ronneberger2015unetconvolutionalnetworksbiomedical} model against our zero-shot methods based on the size of the training dataset. Each U-Net \citep{ronneberger2015unetconvolutionalnetworksbiomedical} model was trained independently from scratch, using randomly sampled subsets of increasing size from each dataset. Training was performed using the standard Adam optimizer with a learning rate of $10^{-3}$ and a batch size of 8, for a maximum of 100 epochs. We applied early stopping based on the loss, with a patience of 5 epochs. Images were pre-processed according to the details provided in Section \ref{preprocessing}. For each training size, we did the same training a 100 times with other random subset to compute confidence intervals. The results indicated that, on average and using this basic U-Net architecture, at least 31 annotated samples are required to outperform our zero-shot method on the Phenobench dataset. For the other datasets, similar results were obtained, except for the Apple Tree dataset, which proved more challenging for a simple U-Net. This dataset contains only 120 samples with relatively low variability, which may explain why a basic supervised model was unable to outperform the zero-shot approach. However, one limitation of such supervised models is their poor ability to generalize, they often struggle to perform well on data significantly different from their training set. In Table \ref{table:crossval-unets}, we evaluate four U-Net models by training each model on one of our selected datasets and testing its performance on the other three datasets. The zero-shot proposed method demonstrate higher robustness toward target domain change.

\begin{table}[ht]
\centering
	\resizebox{\textwidth}{!}{
    \begin{tabular}{lccccc}
    \toprule
        \textbf{Train Dataset / Test Dataset} & \textbf{Phenobench} & \textbf{AppleTreeDataset} & \textbf{Plant Growth} & \textbf{CVPPP2017}\\
        \midrule
        \textbf{Phenobench} & $0.805 \pm 0.128$ & $0.628 \pm 0.094$ & $0.406 \pm 0.197$ & $0.623 \pm 0.126$ \\
        \textbf{AppleTreeDataset} & $0.000 \pm 0.000$ & $0.779 \pm 0.077$ & $0.008 \pm 0.018$ & $0.002 \pm 0.008$ \\
        \textbf{Plant Growth} & $0.000 \pm 0.000$ & $0.410 \pm 0.093$ & $0.925 \pm 0.046$ & $0.442 \pm 0.209$ \\
        \textbf{CVPPP2017} & $0.085 \pm 0.076$ & $0.055 \pm 0.036$ & $0.070 \pm 0.067$ & $0.835 \pm 0.105$ \\
        \bottomrule
    \end{tabular}
    }
\vspace{0.5cm}
\caption{Cross validation of training a Unet on a dataset and testing is on another dataset.}
\label{table:crossval-unets}
\end{table}

\section{Discussion}

Our experiments demonstrate that using Plantnet’s domain-specific features substantially improves zero-shot plant segmentation, with IoU gains of up to 60–70\% over baseline DinoV2 on both sparse (Phenobench) and dense (AppleTreeDataset) dataset.

A key insight is that simple bounding-box prompts for SAM, derived from PCA-thresholded token features, often suffice for high-quality masks, making additional coarse masks useless or even detrimental in most cases. However, highly dense scenes, such as overlapping apple trees, hint that combining box prompts with coarse masks can still help separate subtle features.

Compared to a supervised U-Net, we observed that the number of labeled samples required to outperform our zero-shot approach depends on the difficulty. On the most difficult data set, the AppleTreeDataset, zero-shot outperformed an equivalently small supervised model, illustrating how learned plant-specific representations are valuable when annotated data are scarce or field conditions are diverse.

Potential improvements include refining the PCA-based token segmentation to capture complex plant structures and addressing domain shifts for unusual crop varieties or environments. Additionnaly, extending Plantnet-fine-tuned DinoV2 with class-specific prompts or large language-image models could facilitate more nuanced plant recognition.

\section{Conclusion}


In this work, we introduced a zero-shot segmentation framework leveraging Plantnet’s specialized plant representations, originally developed for species classification, to enable effective plant segmentation in agricultural imagery. By projecting DinoV2-Plantnet features into a principal component space and thresholding the primary component, we generated coarse plant masks, which were then refined by the Segment Anything Model (SAM). Through experiments on four openly available datasets ranging from sparse, top-view sugar beets (Phenobench) to dense apple orchard imagery (AppleTreeDataset), our approach consistently surpassed the baseline DinoV2-based pipeline. Furthermore, our ablation studies revealed that simple box prompts already gives strong performance, while supplying an additional mask to SAM generally does not improve results, except in very dense scenarios.

We also compared our zero-shot method to a supervised baseline (U-Net) on Phenobench and the AppleTreeDataset. Although the U-Net can match or exceed our proposed approach on simpler tasks if given sufficient annotated data (about 30 samples in the Phenobench case), it struggles under limited training data conditions in the AppleTreeDataset and is not able to generalized well. These findings highlight the utility of adding domain specific representations to reduce annotation overhead and improve performance in challenging scenarios.

\section{Acknowledgements}

This work was granted access to the HPC resources of IDRIS under the allocation 2024-AD010115553 made by GENCI.






\newpage

\bibliographystyle{elsarticle-num-names} 
\bibliography{main}



\end{document}